\documentclass[11pt]{article}

\usepackage[preprint]{acl}

\usepackage{times}
\usepackage{latexsym}

\usepackage[T1]{fontenc}

\usepackage[utf8]{inputenc}

\usepackage{microtype}

\usepackage{inconsolata}

\usepackage{graphicx}
\usepackage{subcaption}
\graphicspath{{figures/}{CoT/final/figures/}}
\usepackage{amsmath}
\usepackage{amssymb}
\usepackage{booktabs}
\usepackage{multirow}
\usepackage[capitalize]{cleveref}
\usepackage{nicefrac}
\usepackage{listings}

\lstdefinestyle{promptstyle}{
  basicstyle=\ttfamily\scriptsize,
  breaklines=true,
  breakatwhitespace=false,
  columns=fullflexible,
  frame=single,
  xleftmargin=0pt,
  xrightmargin=0pt,
  framesep=4pt,
  showstringspaces=false,
  keepspaces=true,
  breakindent=0pt,
  postbreak={}
}

\title{Where Do CoT Training Gains Land in LLM based Agents?}

\author{%
\textbf{Jingyu Liu}\textsuperscript{1},
\textbf{Zhiwen Wang}\textsuperscript{2}, 
\textbf{Yuxin Jing}\textsuperscript{2}, 
\textbf{huanyu Zhou}\textsuperscript{2}, 
\\ 
\textbf{Yong Liu}\textsuperscript{1,4,5}$^{\dag}$\\
$^1$ Gaoling School of Artificial Intelligence Renmin University of China, Beijing, China \\
$^2$ ByteDance \\
$^4$ Beijing Key Laboratory of Research on Large Models and Intelligent Governance \\
$^5$ Engineering Research Center of Next-Generation Intelligent Search and Recommendation, MOE~ \\
\tt\footnotesize liujy1016@ruc.edu.cn\\
}

\begin{document}
\maketitle
\begin{abstract}

Chain-of-thought (CoT) reasoning is widely used in language-model
agents, but prior work has shown that verbalized CoT is not always
faithful and may instead reflect post-hoc reasoning, which means
the model already knows the answer before reasoning. We
therefore ask what CoT training is actually improving: is the model
getting better at changing its action through generated reasoning, or
is it getting better at predicting the action directly from the
prompt? We study this question by comparing \emph{prompt actions}
(predicting action without CoT) with
\emph{CoT actions} (predicting action with CoT). Across
checkpoints, prompt-action quality improves
substantially. While interacting with the environment, the relative advantage of CoT actions over prompt
actions remains similar, showing
that CoT training does not widen the advantage of CoT reasoning, and it helps to improve the quality
of prompt actions.
We further find that later checkpoints are less likely to revise the
action in response to CoT, suggesting greater reliance on the prompt.
Motivated by these
patterns, we selectively mask action-token supervision on a fraction
of training examples. This intervention improves out-of-domain
generalization.
\end{abstract}

\section{Introduction}

Large language models are increasingly deployed as interactive agents
\citep{agent_first,scienceworld,world_device_control_agent,CodeAgent,empathetic_agents,gui_agent}.
In many of these systems, the model first generates a chain of thought
and then emits the next action
\citep{deepseek_r1,qwen,gemini}. These agents often benefit from
generating reasoning traces before acting, but prior work shows that
verbalized CoT is not always faithful to the computation that
determines the final output
\citep{FRODO,faithfulness_survey,faithfulness_CoT_benchmark,measure_fithfulness_CoT}.
If the model already has a strong prompt-conditioned guess about what
to do, the reasoning trace may mainly justify that guess rather than
revise it. In long-context agent settings, the prompt itself contains
task instructions, interaction history, and environment feedback that
often narrow down the next action, making such a shortcut especially
plausible.

This raises a training question: when we supervise agents with CoT,
what is actually improving? One possibility is that training mainly
strengthens reasoning-based revision, so the model becomes better at
changing its action after generating a reasoning trace. Another is
that training also strengthens direct action prediction from the
prompt. To distinguish these possibilities, we compare two decoding
modes throughout the paper: \emph{prompt actions}, which are predicted
directly from the prompt without generating CoT, and \emph{CoT
actions}, which are predicted after generating a step-by-step reasoning
trace.

Better performance after CoT generation does not by itself
show that the generated reasoning has become more effective at
changing the action. It may also reflect a model that has become
better at predicting the action directly from the prompt. In other
words, final task performance alone cannot tell us whether training is
increasing the contribution of generated reasoning or simply making the
final action more predictable from information that is already present
in the prompt. This is why we need a more fine-grained diagnosis than
overall success rate alone.

Our analysis combines offline step-level comparisons, online
evaluation on unseen tasks, and conflicting-trace tests, allowing us
to track both how action prediction changes during training and which
source of information exerts more control over the final action at test
time. We complement these behavioral results with attention and
gradient analyses that help explain why the prompt may have a
structural advantage during action prediction, and we use this
diagnosis to motivate a simple reduced action-supervision
intervention.

We find that prompt-action quality improves substantially during
training. Although CoT reasoning still helps, the relative advantage
of CoT actions over prompt actions does not widen as training
progresses. This suggests that CoT training improves performance not
only through better CoT-based revision, but also through stronger
direct action prediction from the prompt. We further find that when
the reasoning trace conflicts with the prompt, later checkpoints are
harder to overturn with conflicting traces, suggesting that the final
action becomes increasingly anchored to the prompt.

Although training requires the model to produce a reasoning trace
before the final action, the model still becomes better at predicting
the action directly from the prompt. To better understand why, we find
that in agent settings the prompt is often much longer than the
reasoning trace and accounts for most of the attention and gradient
mass. As a result, supervision on action tokens can disproportionately
strengthen prompt-based action prediction.

Motivated by this diagnosis,
we selectively reduce action-token supervision on a subset of training
examples and find that this intervention improves OOD
generalization.

Our contributions are: (1)~a diagnostic framework that compares prompt
actions with CoT actions to measure how well the final action can be
predicted from the prompt alone; (2)~behavioral and mechanistic
evidence that CoT supervision can improve direct action prediction
from the prompt in long-context agents; (3)~reduced action
supervision, a training intervention that weakens direct final-action
reinforcement on part of the data and improves OOD
generalization.

\section{Related Work}


\paragraph{Faithfulness of chain-of-thought reasoning.}
Recent work shows that verbalized CoT is not always tightly coupled to
the computation that determines a model's output
\citep{faithfulness_survey,faithfulness_CoT_benchmark,measure_fithfulness_CoT}.
\citet{turpin2023language} show that CoT explanations can omit the
features that actually drive a decision, while
\citet{arcuschin2025chain} find similar unfaithful reasoning
patterns on natural prompts, including cases of implicit post-hoc
rationalization. \citet{FRODO} propose separating reasoning and answer
generation models to improve faithfulness, and
\citet{finetuning_impact_CoT} study whether answer-only fine-tuning
preserves or degrades CoT quality and faithfulness. We differ from
this line of work in two ways: we study the model's training dynamics,
and we focus on reasoning data rather than answer-only training data
in long-context agent settings, where the prompt itself is a rich
source of action-predictive information.

\paragraph{Agent reasoning and generalization.}
A central theme in agent research is the gap between strong in-domain
performance and weak out-of-domain generalization. Prior work reports
mixed conclusions on whether reinforcement learning closes this gap.
\citet{SFT_memorize_RL_generalize,gradient_coupling} argue that RL generalizes better
than supervised fine-tuning, whereas \citet{rl_no_gene} find only
limited gains. Other studies aim to improve generalization through
self-reflection and future prediction \citep{early_experimence},
online data collection and iterative fine-tuning \citep{agentevolver},
meta-reasoning prompts \citep{RLVMR}, or multi-agent interaction
\citep{multi_agent_rl,rema_multi_agent}. Recent critical evaluations
of ReAct-style prompting further suggest that some apparent gains may
depend heavily on prompt construction and exemplar-query similarity
rather than on interleaved reasoning itself
\citep{verma2024brittle,bhambri2025think}. We complement this
literature by relating agent generalization to shortcut learning: in
long-context prompts, the prompt itself may provide a strong shortcut
to the next action, so improved performance need not imply a stronger
reasoning-mediated revision process.


\begin{figure*}[t]
\centering
\includegraphics[width=0.95\textwidth]{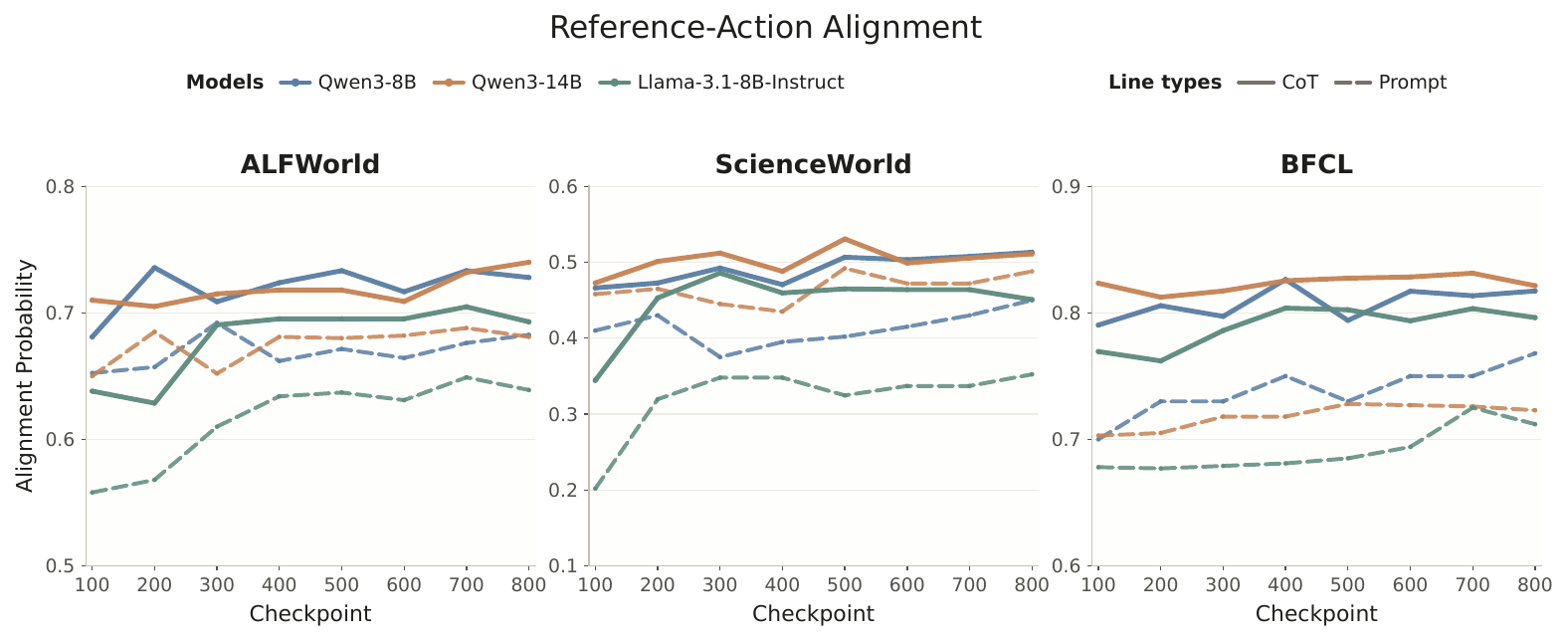}
\caption{Alignment between prompt/CoT actions and the reference action.
On the validation set, both prompt-action and CoT-action accuracies rise
in parallel. This indicates that training gains are increasingly
predictable from the prompt alone, while the CoT-versus-prompt gap
remains flat.}\label{fig:train_reference_alignment_first_cot_prompt_hit_point_plot_nature_refined}
\end{figure*}

\section{Setup and Diagnostic Framework}

We study next-action prediction in long-context agent environments
where the prompt $p$ contains a task description, recent observations,
and interaction history. The model generates a response consisting of
a reasoning trace followed by a final action. We conduct experiments
on three environments: ALFWorld \citep{alfworld} for embodied
household tasks, ScienceWorld \citep{scienceworld} for interactive
scientific reasoning, and BFCL \citep{BFCL} for function-calling and
tool-use evaluation. We train models using both supervised fine-tuning and reinforcement learning.

We define two decoding modes. The \emph{CoT action} is obtained by
instructing the model to generate step-by-step reasoning before
emitting an action. The \emph{prompt action} is obtained by prefilling
the response with the \verb|<action>| tag, which forces the model to
generate the action directly without producing any intermediate
reasoning. Because the prompt action bypasses the reasoning trace
entirely, the gap between prompt-action quality and CoT-action quality
measures how much of the final decision depends on reasoning versus
how much is already recoverable from the prompt alone.

This distinction supports a three-part diagnosis:
\begin{enumerate}
\item We track whether training improves prompt-action quality
  alongside CoT-action quality (\S\ref{sec:training_dynamics},
  \S\ref{sec:matched_context}).
\item We perturb the reasoning trace with a conflicting one and test
  whether the model preserves the original prompt-implied action or
  follows the substituted trace (\S\ref{sec:causal_direction}).
\item We use attention and gradient analyses to explain why the
  prompt can receive more action-time signal than the generated CoT
  (\S\ref{sec:why}).
\end{enumerate}

\section{Training Makes Actions More Predictable from the Prompt}

We first ask whether CoT training also improves direct action
prediction from the prompt alone.
The key question in this section is not whether CoT helps at all, but
where the observed training gains are expressed. If training mainly
improves reasoning-based action revision, then the advantage of CoT
actions over prompt actions should widen over time. If training also
strengthens a direct prompt-to-action pathway, then prompt-action
quality should improve in parallel, and the CoT-versus-prompt gap may
remain relatively stable. We therefore examine this question in three
steps: training dynamics, online evaluation on unseen tasks, and
conflicting-trace
tests.

\subsection{Prompt and CoT Action Accuracy Improve in Parallel}\label{sec:training_dynamics}

During supervised fine-tuning, we split the data into a training set
and a validation set.
We train the model on the training set and then evaluate whether prompt
actions and CoT actions match the reference action on the validation
set, testing whether the model has learned to predict the action
directly from the prompt.
As shown in
Figure~\ref{fig:train_reference_alignment_first_cot_prompt_hit_point_plot_nature_refined},
these two accuracies rise in parallel over training: when the model
gets better at producing the correct final action after generating
CoT, it also gets better at predicting the action directly from the
prompt.
This parallel trend is the first signal that the gains from CoT
training are not expressed only after the reasoning trace is produced.
Instead, training appears to improve how much of the final action is
already recoverable from the prompt itself. In other words, the model
does not merely become better at arriving at the right action after
reasoning; it also becomes better at predicting that action before
any reasoning is generated.

\begin{figure*}[t]
\centering
\includegraphics[width=0.95\textwidth]{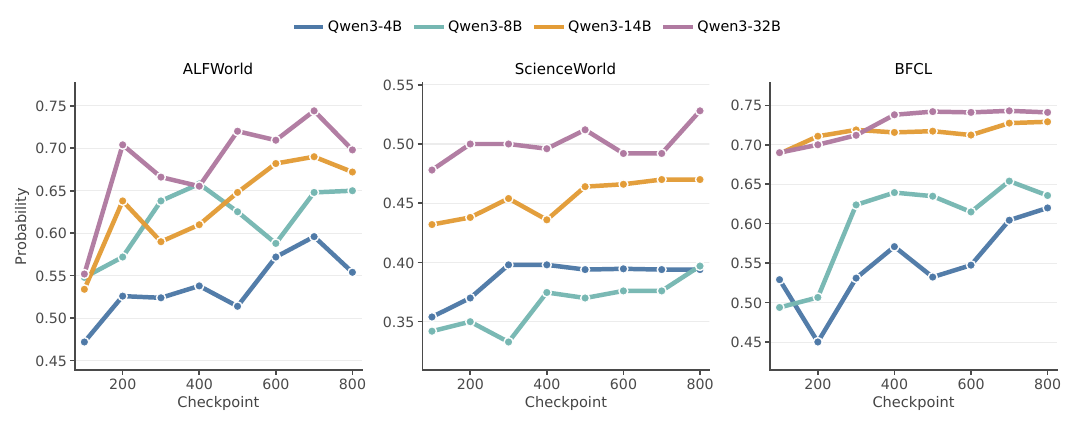}
\caption{Prompt/CoT-action consistency during training.
Each panel corresponds to one environment;
colors denote model size. Across environments, higher-performing checkpoints also
tend to exhibit higher prompt/CoT consistency.}\label{fig:truncated_cot_majority_consistency_sft_in_domain_nature}
\end{figure*}

We also evaluate consistency between prompt actions and CoT actions.
As shown in Figure~\ref{fig:truncated_cot_majority_consistency_sft_in_domain_nature},
the prompt/CoT-consistency curves generally rise over training.
The model therefore reaches the same action more often with and
without the full reasoning trace.
This rising consistency sharpens the interpretation of the accuracy
trends above. The important pattern is not only that both decoding
modes improve, but that they converge more often to the same decision.
As training proceeds, the final action increasingly looks like
something that can already be inferred from the prompt,
with the reasoning trace serving more as a continuation of that
decision process than as a separate source of revision.

\begin{figure*}[htbp]
    \centering
    \begin{subfigure}{0.48\textwidth}
        \centering
          \includegraphics[width=\linewidth]{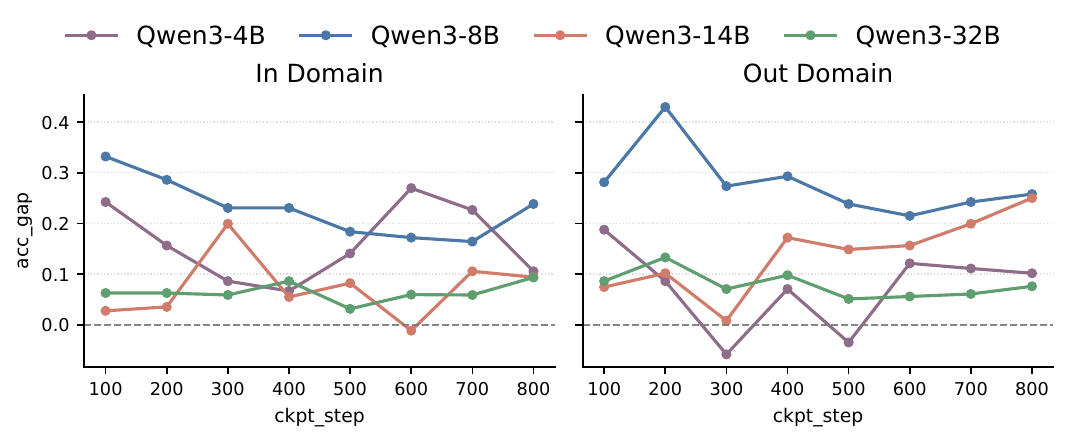}
          \caption{Trajectory success gap between CoT-action and
          prompt-action execution on ALFWorld}\label{fig:alf_acc_gap}
    \end{subfigure}
    \hfill
    \begin{subfigure}{0.48\textwidth}
        \centering
      \includegraphics[width=1\linewidth]{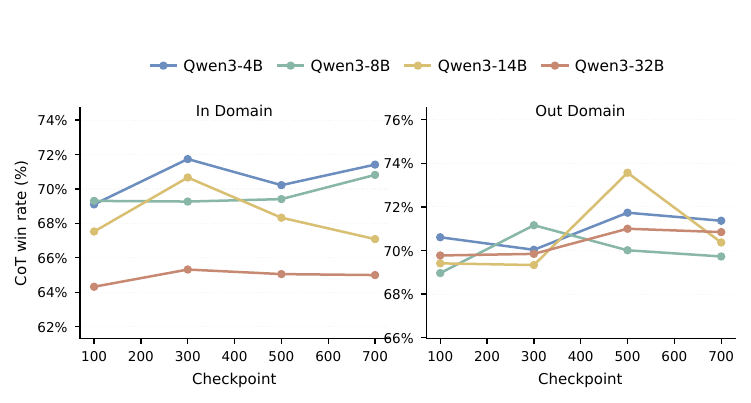}
      \caption{GPT-5.4 preference rate for CoT actions over prompt
      actions in matched decision contexts on ALFWorld}\label{fig:alf_CoT_winrate}
    \end{subfigure}
    \caption{Online evaluation of how well actions can be predicted
    from the prompt on unseen tasks. Panel (a)
    compares trajectory success under prompt versus CoT actions, and
    panel (b) compares local action quality in the same decision
    context using GPT-5.4 judgments. Both views show that the
    CoT-versus-prompt gap stays largely flat across checkpoints.}\label{fig:combined}
\end{figure*}

This section establishes the basic behavioral pattern: training
improves CoT-conditioned action generation and prompt-only action
prediction together. We return in \S\ref{sec:why} to why action
supervision may favor prompt tokens in these long-context settings.
At a high level, Figure~\ref{fig:train_reference_alignment_first_cot_prompt_hit_point_plot_nature_refined}
and Figure~\ref{fig:truncated_cot_majority_consistency_sft_in_domain_nature}
tell a consistent story. Better checkpoints are not only more accurate
after generating CoT; they are also better at recovering the same
action directly from the prompt, and more likely to agree across the
two decoding modes. This makes direct action prediction from the prompt
an increasingly important part of the model's observed behavior over
training.

\subsection{Direct Prompt-vs-CoT Comparisons in Online Evaluation}\label{sec:matched_context}

We next test whether the same pattern persists during online
evaluation on unseen tasks.

We split the tasks into in-domain
train/test sets and OOD test sets; the detailed split is shown in the
Appendix. This test matters because training-set predictability from
the prompt alone
could reflect memorization or distribution-specific coupling between
prompt and action.

First, we evaluate performance when the agent interacts with the
environment using the prompt action rather than the CoT action. As
shown in Figure~\ref{fig:alf_acc_gap},
the gap between the two
decoding modes remains flat over training. We show more results in the Appendix.
This shows that the extra benefit
of generating CoT before acting does not expand with training. The
improvement therefore does not appear primarily as a growing
CoT-specific advantage; instead, a substantial part of it remains
recoverable from the prompt alone even on unseen tasks.

\begin{figure*}[t]
\centering
\includegraphics[width=0.95\textwidth]{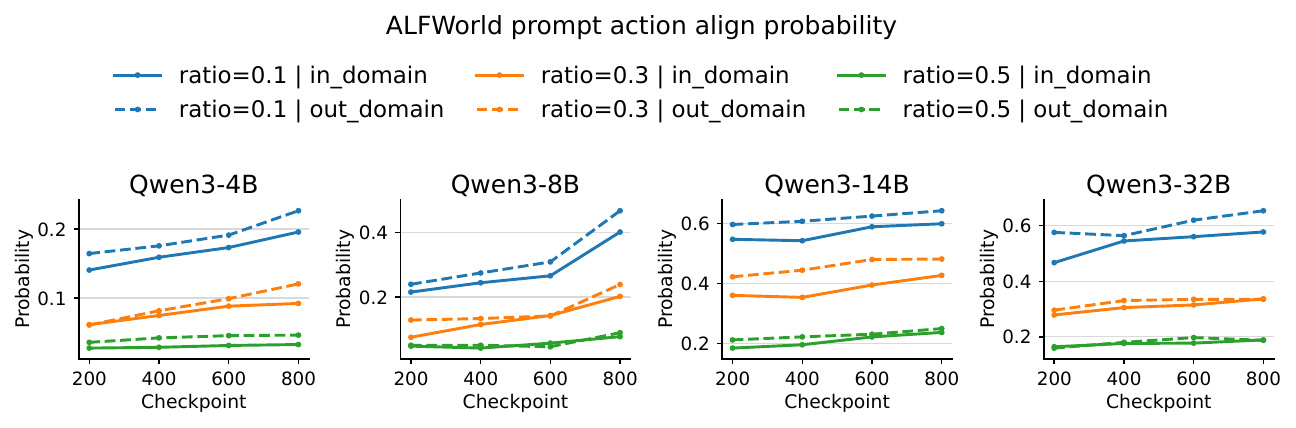}
\caption{Checkpoint-wise consistency under perturbed reasoning traces
in ALFWorld. Each panel corresponds to one model size; colors denote
the truncation ratio of the substituted reasoning trace, and solid
versus dashed lines distinguish in-domain and out-of-domain settings.
Later checkpoints more often preserve the original prompt-based action
under conflicting traces, especially for shorter truncations,
consistent with stronger reliance on the prompt.}\label{fig:alf_perturbed_truncate_consistency_ckpt}
\end{figure*}

Second, we compare prompt actions and CoT actions in the same decision
context by asking GPT-5.4 which action is better.
Figure~\ref{fig:alf_CoT_winrate} shows that the win rate of CoT
actions over prompt actions is stable across checkpoints. This
action-level comparison reduces the attribution problem in
trajectory-level outcomes: even if later recovery or
mistakes make full-trajectory success noisy, the judge still asks
whether the generated CoT yields a locally better action than the
prompt-based action at the current decision point.
We also manually checked GPT-5.4's judgments on 100 examples, and they
aligned with human judgments in 93\% of cases.
For OOD tasks, CoT-revised actions likewise show only limited
improvement over prompt actions.
Together with the trajectory-level evaluation, this action-level view
shows that the flat-gap pattern is not simply an artifact of long-horizon
execution noise. Even when we compare the two candidate actions in the
same local decision context, later checkpoints do not show a widening
advantage for CoT-conditioned generation. The main training trend is
therefore one of parallel improvement rather than increasing separation
between the two pathways.

CoT actions retain a consistent advantage over prompt actions
throughout training, so reasoning still helps. However, that advantage
does not grow: both trajectory-level execution and local action-level
judgments show the same flat-gap pattern in online evaluation on
unseen tasks.
This combination of results is central for the rest of the paper.
CoT remains useful, but its relative benefit stays roughly constant
even as overall performance rises. The natural reading is that training
is making more of the correct action predictable from the prompt,
rather than increasingly shifting the final decision through generated
reasoning alone.

\subsection{Does Final Generation Increasingly Rely on the Shortcut?}\label{sec:causal_direction}

The previous sections compare prompt and CoT actions under standard
decoding. We now use a stronger probe: when the supplied reasoning
trace conflicts with the prompt, which one does the model follow?

To test this, we follow prior work
\citep{faithfulness_CoT_benchmark,measure_fithfulness_CoT} and replace
the original reasoning trace for a prompt with a randomly chosen trace
that supports a different final action. We then truncate the inserted
trace at different ratios and ask whether the resulting action agrees
with the original prompt-based action or with the action implied by the
substituted trace. Figure~\ref{fig:alf_perturbed_truncate_consistency_ckpt}
shows the result in ALFWorld: across checkpoints, later models
preserve the original prompt-based action more often, even under
conflicting traces. This broad upward trend, replicated across BFCL and
ScienceWorld (Appendix~\ref{sec:increasing_prompt_side_reliance}),
shows that later checkpoints are harder to overturn with substituted
reasoning traces. Similar checkpoint-wise trends under reinforcement
learning and for the Llama model are reported in
Appendix Figures~\ref{fig:grpo_all_samples_ckpt_nature}
and~\ref{fig:all_samples_ckpt_nature}. A related pattern is that
out-of-domain settings are often more prompt-anchored than in-domain
ones, which is consistent with the view that when reasoning is less
reliable or harder to use, the model falls back more heavily on the
prompt-based pathway.
Appendix~\ref{sec:increasing_prompt_side_reliance} shows more results.

\section{Why the Prompt Has a Structural Advantage}\label{sec:why}

\begin{figure}[t]
\centering
\includegraphics[width=0.45\textwidth]{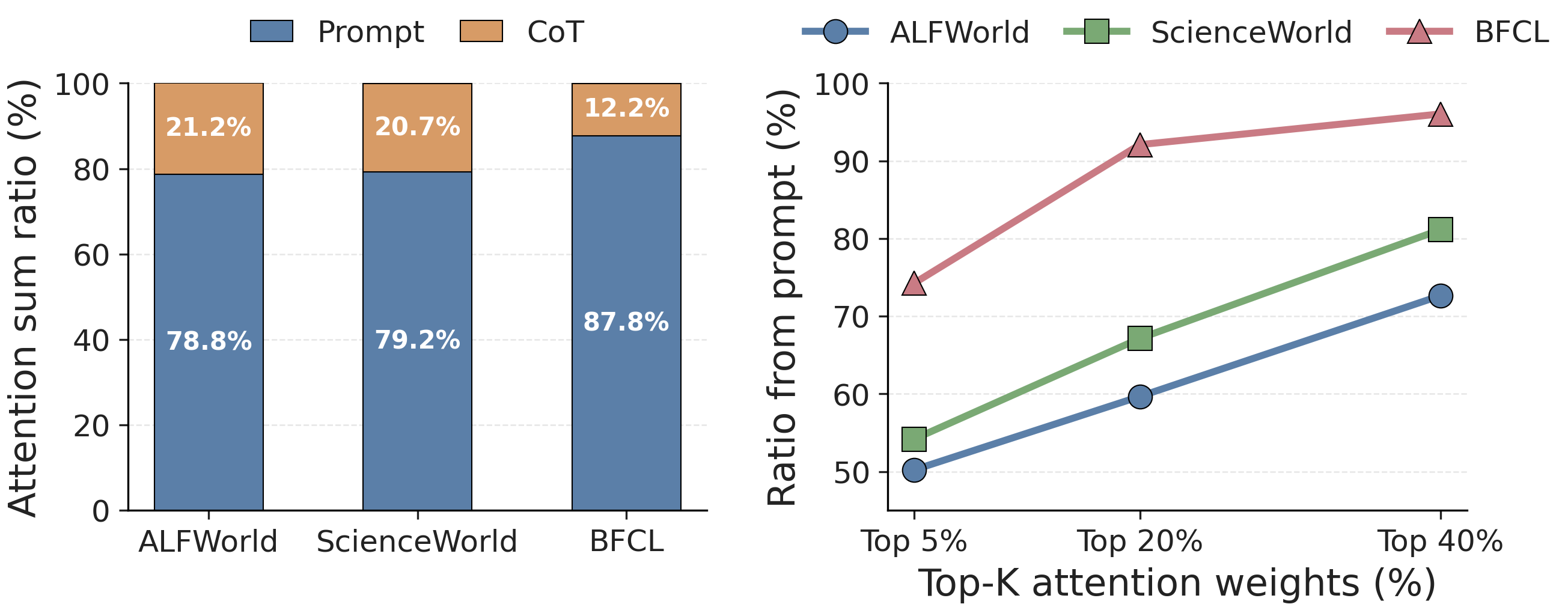}
\caption{Prompt versus CoT attention during action
prediction. Left: total attention mass assigned to prompt and CoT
tokens across environments. Right: the fraction of the top-$K$\%
highest-attention positions that come from prompt tokens. Both views
show that prompt tokens dominate action-time attention. We average
attention across all layers and heads.}\label{fig:attention_analysis_visualization}
\end{figure}

We next ask why action supervision in long-context agents may favor
prompt tokens over generated CoT tokens.

In autoregressive generation, action tokens are predicted from the
full available context, which includes both the prompt and the
previously generated rationale. In agent settings, however, the prompt
is often substantially longer than the CoT because it contains task
instructions, interaction history, and current environment feedback.
This creates a structural asymmetry: when the model generates the
final action, information from the prompt can occupy a larger share of the
available evidence stream.

Figure~\ref{fig:attention_analysis_visualization} also supports this
asymmetry. Nearly 80\% of the attention mass during action generation
falls on prompt tokens rather than CoT tokens. While part of this
effect is mechanical (the prompt is typically much longer than the
CoT), prompt tokens still dominate even among the highest-attention
positions, where more than half come from the prompt. Action
prediction is therefore empirically prompt-dominated in these
settings, beyond what prompt length alone would
imply.

Our goal here is to measure aggregate pathway dominance at action
prediction time, not per-token informativeness. We therefore report
the total attention mass and gradient share assigned to prompt tokens
versus CoT tokens, rather than length-normalized averages. In
agent settings, prompt length is itself part of the structural
asymmetry: a longer prompt can cause the prompt-based pathway to
receive a larger share of the action-supervision signal.

Along the value path of self-attention, the gradient contribution from
context token $i$ to action-token supervision is weighted by its
attention weight. In particular, if $\mathcal{L}_t$ denotes the loss on
action position $t$ and $o_t$ is the corresponding attention output,
then
\begin{equation}
\frac{\partial \mathcal{L}_t}{\partial v_i}
= \alpha_{t,i}\,\delta_t,
\qquad
\delta_t = \frac{\partial \mathcal{L}_t}{\partial o_t}.
\end{equation}
Thus, if prompt tokens consistently receive a larger share of the
attention mass during action generation, they also receive a larger
share of the optimization signal along this pathway. A full derivation
is provided in Appendix~\ref{sec:value_path_derivation}. This implies
an optimization bias along the value path: when prompt features are
already predictive of the next action, gradient descent may
preferentially strengthen the direct prompt-to-action pathway,
leaving the rationale to function as a weaker refinement signal
rather than the main route to the decision.

To complement this forward-pass view, we also compute a direct
gradient-based diagnostic of where the action-supervision signal flows
during backpropagation. As shown in
Table~\ref{tab:per_token_gradient_norm}, prompt tokens receive a larger
share of the gradient signal than CoT tokens.

\begin{figure*}[t]
\centering
\includegraphics[width=0.95\textwidth]{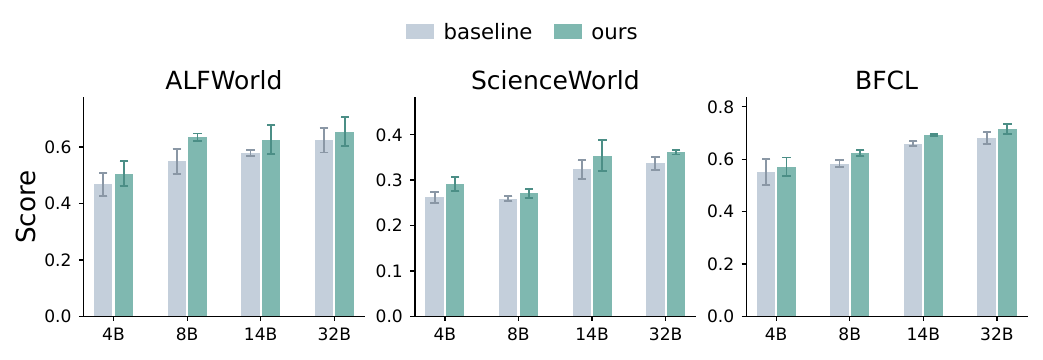}
\caption{Mean evaluation score on OOD tasks under vanilla CoT supervision
(\textit{baseline}) and reduced action supervision (\textit{ours})
across environments and model sizes. Reduced action supervision
improves performance in most environment-model combinations, with the
largest gains appearing in ALFWorld and ScienceWorld.}\label{fig:per_cot_ood_nature}
\end{figure*}

Together, the attention and gradient evidence provide a plausible
optimization-level explanation for the behavioral patterns above.
Agent prompts are substantially longer than the reasoning traces they
surround, and prompt tokens receive the dominant share of attention
and gradient signal during action prediction. We also conduct
experiments on relatively short-prompt datasets, where
prompt/CoT-action consistency does not increase much over training
(Figure~\ref{fig:answer_consistency_three_task_plot}).

\section{A Training Intervention Motivated by These Patterns}\label{sec:intervention}

If action-token supervision mainly reinforces direct action prediction
from the prompt, then reducing that supervision on part of the training data
should weaken this effect.

We therefore remove action supervision on a subset of training
examples.
Concretely, for a randomly selected fraction $k\%$ of training samples,
we mask the loss on the final action tokens and optimize only the CoT
span. For the remaining samples, we keep the standard supervision
unchanged. Let $M(t)$ be an indicator that equals $1$ when token
position $t$ belongs to the CoT span and $0$ otherwise. For masked
examples, we train with
\begin{equation}
\mathcal{L}_{\mathrm{cot}} = - \sum_{t=1}^{|y|} M(t)\log p_{\theta}(y_t \mid p, y_{<t}).
\end{equation}
This objective removes the direct optimization pressure on final
action tokens on the selected samples. Because the remaining samples
still preserve standard action supervision, the model continues to
receive explicit training signals for action generation. We also keep
supervision on the \verb|<action>| tag so that the model still learns
when to transition from reasoning to action generation. Reduced action
supervision therefore does not remove reasoning supervision; it
selectively weakens how often the final action loss can directly
reward shortcut learning from the prompt.

We also apply this intervention to reinforcement learning. There,
we randomly select $k\%$ of training samples, mask the loss on
the final action tokens, and optimize only the CoT span,
while keeping the advantage estimates unchanged.

\subsection{Evaluation}


\begin{figure*}[t]
\centering
\includegraphics[width=0.95\textwidth]{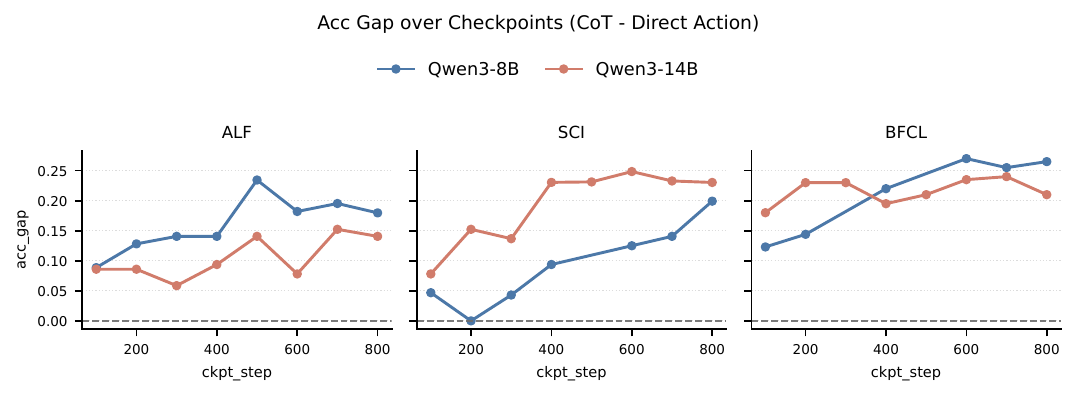}
\caption{CoT-minus-prompt action gap under reduced action supervision
across checkpoints. Each panel corresponds to one environment, and
each line to one model size. Relative to the flatter gaps observed
under vanilla training, larger gaps in several settings suggest more
room for CoT-based action revision, although the effect varies by
environment.}\label{fig:actionmask_acc_gap_compare_plot}
\end{figure*}



This evaluation serves two goals: to test whether reduced action
supervision preserves or improves standard task performance, and to
test whether it changes the prompt-versus-CoT diagnostics in the
predicted direction.

We evaluate the intervention in three agent settings: ALFWorld for
embodied household tasks \citep{alfworld}, ScienceWorld for
interactive scientific reasoning \citep{scienceworld}, and BFCL for
function-calling and tool-use evaluation \citep{BFCL}. Across these
environments, we compare standard CoT-supervised training with reduced
action supervision and summarize checkpoint-wise evidence about
shortcut reliance and rationale sensitivity under our diagnostics.
We train for 5 epochs in SFT and 200 steps in RL, use learning rates
of 1e-5 for SFT and 1e-6 for RL, and run each setting three times,
reporting mean performance. We set $k=0.3$ for both SFT and RL; an
ablation over $k$ is shown in Table~\ref{tab:k_performance}. More
detailed settings are provided in the Appendix.

Figure~\ref{fig:per_cot_ood_nature} shows that reduced action
supervision improves performance in most environment-model
combinations.
Figure~\ref{fig:actionmask_acc_gap_compare_plot}
reports the CoT-minus-prompt action gap under reduced action
supervision. Relative to the flatter gap pattern under vanilla
training, the gap is generally larger in ALFWorld and ScienceWorld,
with more mixed changes in BFCL. We treat this shift as supportive,
though not by itself definitive, evidence that the intervention
reduces reliance on prompt-only action prediction and leaves more room
for CoT-based action revision. The corresponding comparison under GRPO
is shown in Table \ref{tab:per_grpo} and Figure~\ref{fig:grpo_vs_ours_8b_gap_three_task_plot}.

The variation across environments is also informative. Gains are
largest in ALFWorld and ScienceWorld and more modest in BFCL,
suggesting that the intervention helps most when a stronger
prompt-based shortcut would otherwise dominate action
selection. Together with the performance improvements in
Figure~\ref{fig:per_cot_ood_nature}, this pattern is consistent with
the shortcut diagnosis: weakening the prompt-to-action shortcut can
improve task performance while leaving more room for CoT-based
revision.
The more mixed pattern in BFCL is informative rather than purely
negative. Compared with ALFWorld and ScienceWorld, BFCL uses more
structured prompts and a more constrained action space, which may
leave less room for the intervention to change the balance between
prompt-based prediction and CoT-based revision.

\begin{table}[htbp]
  \centering

    \begin{tabular}{cccc}
    \toprule
          & ALFWorld & ScienceWorld & BFCL \\
    \midrule
    SFT   & 0.55  & 0.25  & 0.58 \\
    DPO   & 0.59  & 0.29  & 0.62 \\
    FRODO & 0.61  & 0.30  & 0.64 \\
    SFT+  & 0.63  & 0.27  & 0.62 \\
    DPO+  & 0.65  & 0.35  & 0.67 \\
    \bottomrule
    \end{tabular}%
  \caption{Performance of Qwen3-8B against other baselines.
  SFT+ denotes our method with SFT, and DPO+ denotes our method with DPO.}\label{tab:per_baseline_comparison}
\end{table}%
We also compare our method with other baselines, including DPO
\citep{DPO} and FRODO \citep{FRODO}. And following \citet{FRODO}, we use response randomly selected from other prompt response pairs as the rejected response.
As shown in Table~\ref{tab:per_baseline_comparison}, although our
method with SFT underperforms on some tasks, it performs well when
combined with DPO. It is also worth noting that FRODO
\citep{FRODO} uses DPO for training.

\begin{figure}[t]
\centering
\includegraphics[width=0.45\textwidth]{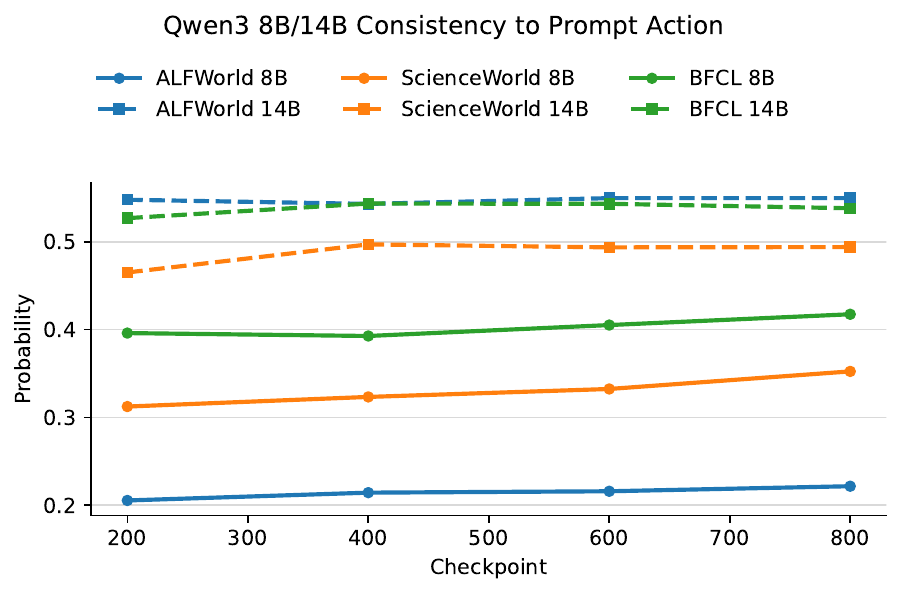}
\caption{Consistency with the prompt action under perturbed reasoning
traces when the truncation ratio is set to $0.1$.}\label{fig:qwen3_8b_14b_three_task_all_samples_ckpt_ratio0.1_in_domain}
\end{figure}

Finally, as shown in Figure~\ref{fig:qwen3_8b_14b_three_task_all_samples_ckpt_ratio0.1_in_domain},
under the intervention, consistency with the prompt action under
perturbed reasoning traces does not increase over training.

This result suggests that our method can effectively mitigate the
prompt-action shortcut, and that the performance gain is mainly driven
by CoT-conditioned action revision, which helps explain the improved
OOD performance.

\section{Conclusion}
We study how CoT training changes the role of direct action prediction
from the prompt in long-context language-model agents by comparing actions
decoded directly from the prompt with actions decoded after a
reasoning trace. We find that, training improves
direct action prediction from the prompt. Later checkpoints are also harder to overturn with
conflicting reasoning traces. These findings do not settle whether
latent reasoning ability improves in an absolute sense, but they do
show that, in the long-context agent settings we study, gains from
standard CoT supervision should not be interpreted as evidence that
CoT-based revision alone has become more important.

\section*{Limitations}

Prompt actions serve only as a
behavioral proxy for how much of the final action can be predicted
from the prompt rather than a direct
measurement of internal computation. In addition, some of our local
action-quality judgments rely on GPT-5.4 as a judge, which may
introduce evaluation bias.

\IfFileExists{../refs.bib}
  {\bibliography{../refs}}
  {\bibliography{CoT/refs}}

\appendix

\noindent This appendix is organized as follows. We first report
experimental details and an ablation for the masking ratio. We then
provide supplementary behavioral evidence for increasing shortcut
dependence during training, followed by additional behavioral
diagnostics and outcome-level analyses. Next, we present the
mechanistic derivation and gradient-based diagnostic that support the
main-text interpretation. We conclude with qualitative materials and
the full prompts used in our experiments.

\section*{LLM usage}
We use LLMs to help polish the paper and draft figures.

\section{Experimental Details}
This section collects the setup information needed to interpret the
results in the main text and the Appendix. We first summarize the data
split and training configuration, and then report the masking-ratio
ablation used in the reduced action-supervision study.

\subsection{Data Split}
We conduct our experiments on three environments: ALFWorld,
ScienceWorld, and BFCL. We use a maximum of 30 interaction steps for
ALFWorld and ScienceWorld, and 15 for BFCL. For ALFWorld, following
\citet{RLVMR}, we designate Cool \& Place and Pick Two \& Place as
unseen tasks for OOD evaluation. For ScienceWorld, the final task
type of each topic is reserved for OOD evaluation. For BFCL, we use
the long-context setting for OOD evaluation.

\subsection{Training Details}
We train for 5 epochs in SFT and 200 steps in RL, using a learning
rate of 1e-5 for SFT and 1e-6 for GRPO \citep{GRPO}. For SFT, we use a batch size of 16. For GRPO,
we use a batch size of 32, and the group size is set to 8. To
penalize outputs that did not adhere to the required format, we apply
a reward penalty of -0.1.
The KL regularization coefficient is set to 0.01.
We use GPT-5.4 to collect trajectories and select successful
trajectories for SFT.

\begin{table*}[htbp]
  \centering
  
    \begin{tabular}{cccccccc}
    \toprule
          &       & \multicolumn{2}{c}{ALFWorld} & \multicolumn{2}{c}{SCIWorld} & \multicolumn{2}{c}{BFCL} \\
          &       & in domain & out domain & in domain & out domain & in domain & out domain \\
    \midrule
    \multirow{2}[2]{*}{Qwen3-4B} & GRPO  & 0.82  & 0.44  & 0.46  & 0.22  & 0.69  & 0.61 \\
          & ours  & 0.83  & 0.52  & 0.49  & 0.24  & 0.71  & 0.64 \\
    \midrule
    \multirow{2}[2]{*}{Qwen3-8B} & GRPO  & 0.95  & 0.86  & 0.57  & 0.4   & 0.74  & 0.63 \\
          & ours  & 0.95  & 0.89  & 0.61  & 0.44  & 0.77  & 0.69 \\
    \bottomrule
    \end{tabular}%
    \caption{The performance when conducting GRPO training.}
  \label{tab:per_grpo}%
\end{table*}%

\subsection{Contrast with short-prompt settings.} \label{sec:short_prompt_settings}
\begin{figure}
\centering
\includegraphics[width=0.45\textwidth]{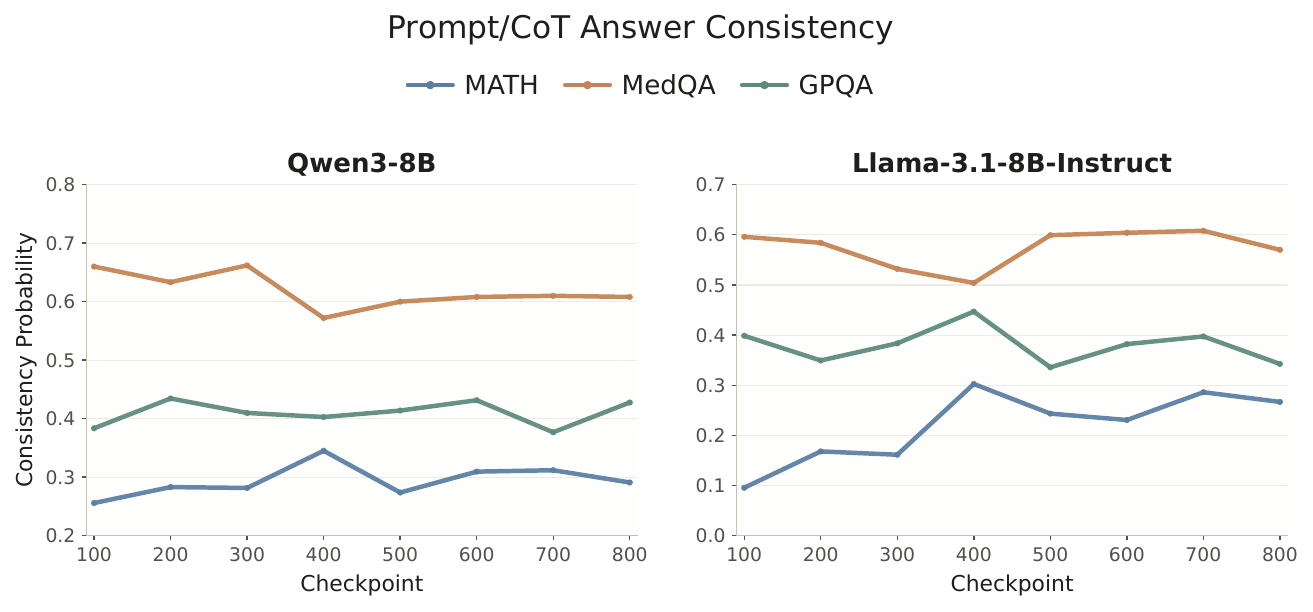}
\caption{Prompt/CoT-action consistency in MATH, MedQA, and GPQA}\label{fig:answer_consistency_three_task_plot}
\end{figure}

To further evaluate our hypothesis, we also consider short-prompt
settings (MATH, MedQA, and GPQA), where prompts are much shorter and
do not carry rich interaction histories. As shown in
Figure~\ref{fig:answer_consistency_three_task_plot}, the consistency
between prompt actions and CoT actions in these short-prompt datasets
does not increase over training, which is also consistent with
previous results \citep{finetuning_impact_CoT,FRODO}. This contrast
supports the view that the structural length asymmetry specific to
agent prompts is a key driver of the shortcut pattern we observe.

\subsection{Masking-Ratio Ablation}
Table~\ref{tab:k_performance} reports the performance of our reduced
action-supervision objective under different masking ratios $k$ for
Qwen3-8B.

\begin{table}[htbp]
  \centering

    \begin{tabular}{cccc}
    \toprule
          & ALFWorld & ScienceWorld & BFCL \\
    \midrule
    0.1   & 0.61  & 0.26  & 0.61  \\
    0.3   & 0.63  & 0.27  & 0.62  \\
    0.5   & 0.61  & 0.24  & 0.59  \\
    0.7   & 0.6  & 0.27  & 0.61  \\
    \bottomrule
    \end{tabular}%
    \caption{Performance under different values of $k$ for Qwen3-8B.}\label{tab:k_performance}
\end{table}%

\section{Conflicting-Trace Evidence for Reliance on the Prompt}\label{sec:increasing_prompt_side_reliance}
We next present supplementary evidence that most directly probes the
paper's directional claim. By replacing the model's original reasoning
trace with a conflicting one, this section asks whether training makes
final actions more or less anchored to the prompt.

The previous analysis suggests that training benefits direct action
prediction from the prompt more strongly than cases that require revising the prompt-based
decision. A natural question is whether this phenomenon also changes
the extent to which the final action is anchored by the prompt relative
to the generated trace. Previous work \citep{FRODO} finds that fine-tuning on
CoT data can increase trace sensitivity in mathematical and QA tasks.
However, our results suggest that, for agentic tasks, training may
instead increase the degree to which final actions remain predictable
from the prompt.

The core diagnostic logic in this section is simple. If training makes
the final action increasingly sensitive to the supplied reasoning
trace, then a conflicting substituted trace should overturn the
original prompt-based action more often as training proceeds. If,
instead, the observable gains are increasingly recoverable from the
prompt, then the model should preserve the original prompt-based action
more often even when the inserted reasoning trace supports a different
answer.


To probe how the model determines the final action, we follow
\citet{faithfulness_CoT_benchmark,measure_fithfulness_CoT} and
perturb the model's reasoning trace while measuring the resulting
action. For a prompt $p_1$ with its original reasoning trace $t_1$, we
replace the reasoning trace with a random trace $t_2$ that leads to a
different final action, and then measure the action generated from the
mixed input $(p_1, t_2)$. We further truncate $t_2$ at different
ratios and track how often the resulting action agrees with the
original prompt-based action versus the action associated with the
substituted trace. We use prefilling so that the model treats the
inserted thinking trace as part of its own response, and then append
the token \verb|<action>| to force direct action prediction without
additional reasoning. Because the substituted trace conflicts with the
original prompt, agreement with the original prompt-based action serves
as a proxy for how strongly the model preserves its prompt-conditioned
decision under conflicting evidence.
This probe goes beyond comparing two decoding modes under normal
generation. The earlier analyses show that prompt actions and CoT
actions improve together; the conflicting-trace test asks which source
of information exerts more control when the two are explicitly put in
tension. It therefore provides a more direct view of how strongly the
final action remains anchored to the prompt.


Figure~\ref{fig:alf_perturbed_truncate_consistency_ckpt} further indicates
that agreement with the original prompt-based action increases over
training checkpoints. This trend is consistent with increasing
prompt anchoring of the final action relative to the influence of
the substituted reasoning trace. This interpretation is especially
natural because, if training systematically improved the relative value
of CoT-based action revision, one would expect the substituted trace to
overturn the original prompt-based action more often, not less, as
training proceeds. The overall pattern is therefore consistent with a
setting in which the rationale has limited leverage over an action that
is already largely determined by the prompt.
This trend also fits naturally with the flat-gap results above. If
later checkpoints already encode a stronger prompt-based guess about the
next action, then conflicting reasoning traces should find it harder to
move the final decision. The behavioral picture is therefore coherent
across analyses: training improves CoT-conditioned performance, but it
also makes the model increasingly likely to preserve the action that is
recoverable from the prompt.

\begin{figure*}[t]
\centering
\includegraphics[width=0.95\textwidth]{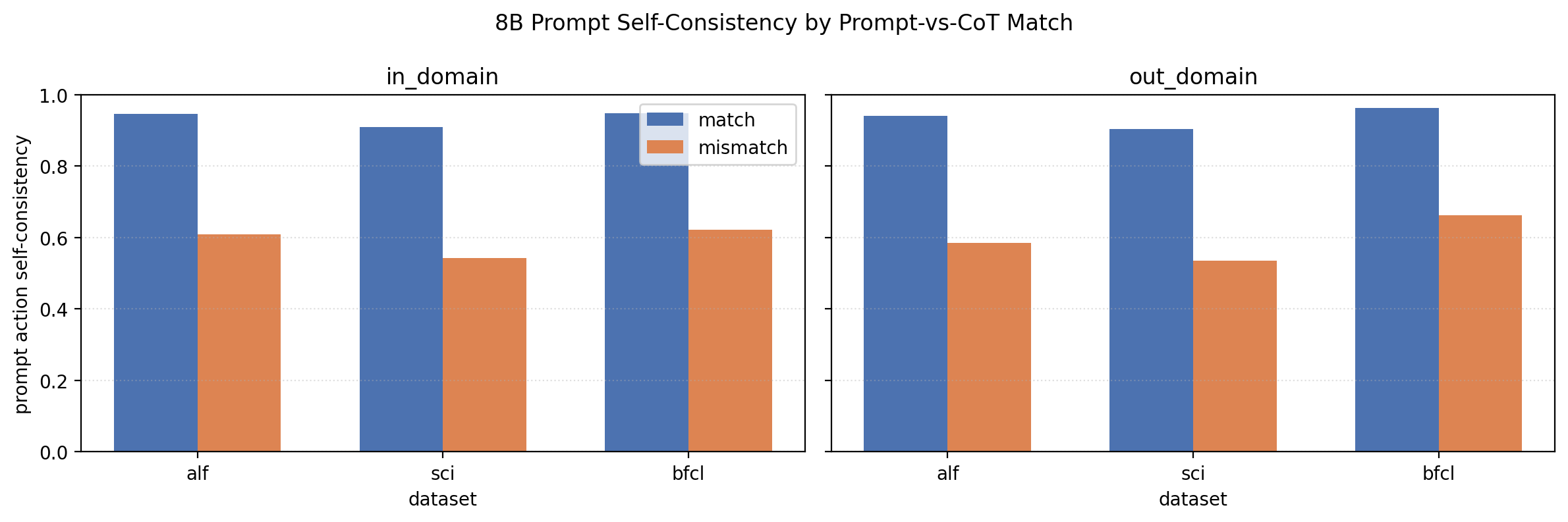}
\caption{Prompt-action self-consistency conditioned on whether the prompt action matches or mismatches the CoT action.}\label{fig:prompt_action_match_self_consistency_ckpt}
\end{figure*}

Figure~\ref{fig:prompt_action_match_self_consistency_ckpt} indicates that
prompt-action self-consistency is positively related to prompt/CoT
action consistency. This pattern suggests that when the model is
more stable in its prompt-based action, the reasoning process has
less room to revise it. When prompt-based behavior is less
stable, reasoning may have more room to change the final
action.

Taken together, these results are consistent with training increasingly
favoring prompt-conditioned action generation under conflicting
evidence, while also making the model more likely to base its decisions
on prompt information. These effects may contribute to stronger overall
performance, but heavier prompt reliance can also encourage
overfitting and lead to weaker out-of-domain generalization. As shown in
Figure~\ref{fig:alf_perturbed_truncate_consistency_ckpt},
out-of-domain tasks exhibit stronger dependence on the prompt than
in-domain tasks. However, Figure~\ref{fig:prompt_action_match_self_consistency_ckpt}
indicates that prompt-action self-consistency on those out-of-domain tasks
is similar to that on in-domain tasks. This pattern suggests that stronger
dependence on the prompt out of domain does not necessarily arise from
greater confidence in the prompt action alone. One possible
interpretation is that reasoning on unfamiliar tasks is of lower
quality, making the CoT less able to revise the prompt-based action and
thereby contributing to poor generalization.


\begin{figure*}[t]
\centering
\includegraphics[width=0.95\textwidth]{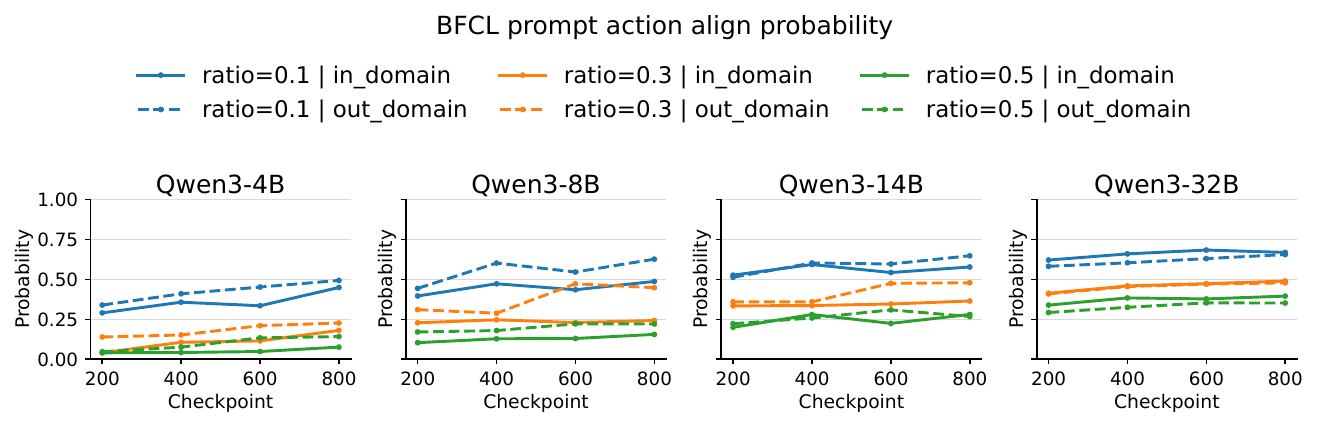}
\caption{Checkpoint-wise consistency under perturbed reasoning traces in BFCL. Increasing agreement with the original prompt-based action is consistent with a stronger prompt-based pathway over training.}\label{fig:bfcl_perturbed_truncate_consistency_ckpt}
\end{figure*}


\begin{figure*}[t]
\centering
\includegraphics[width=0.95\textwidth]{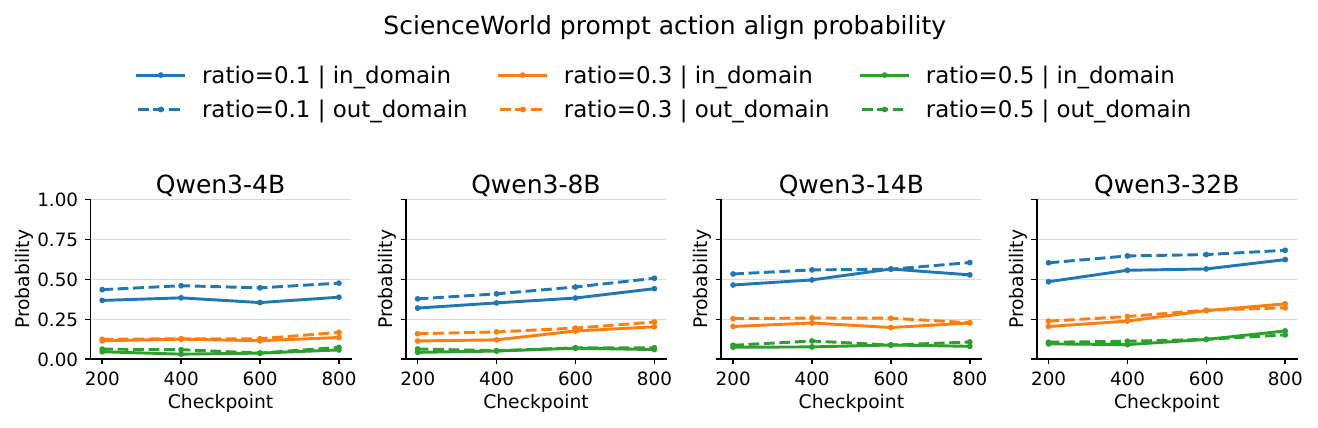}
\caption{Checkpoint-wise consistency under perturbed reasoning traces in ScienceWorld. Increasing agreement with the original prompt-based action is consistent with a stronger prompt-based pathway over training.}\label{fig:sci_perturbed_truncate_consistency_ckpt}
\end{figure*}


\begin{figure*}[t]
\centering
\includegraphics[width=0.95\textwidth]{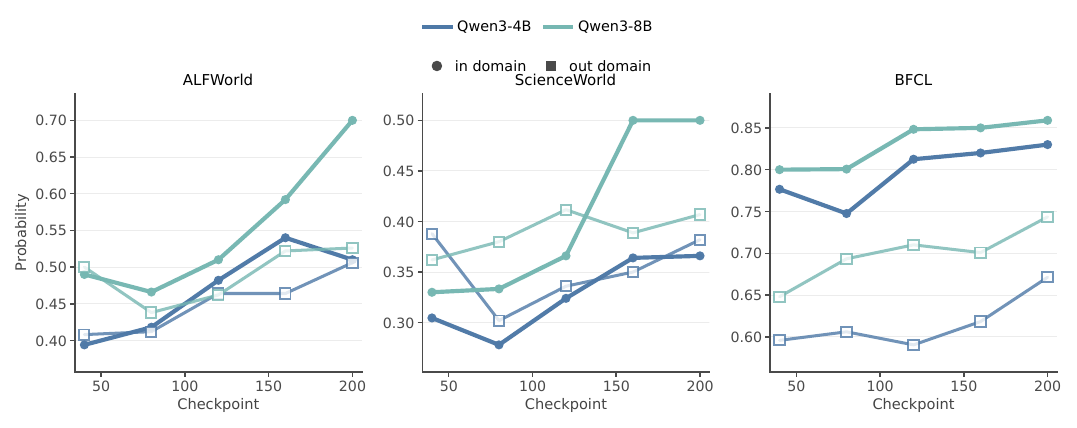}
\caption{Prompt/CoT-action consistency during reinforcement learning.
Each panel corresponds to one environment;
colors denote model size. Across environments, higher-performing checkpoints also
tend to exhibit higher prompt/CoT consistency.}\label{fig:truncated_cot_majority_consistency_rl_combined_consistency_nature}
\end{figure*}

\begin{figure*}[t]
\centering
\includegraphics[width=0.95\textwidth]{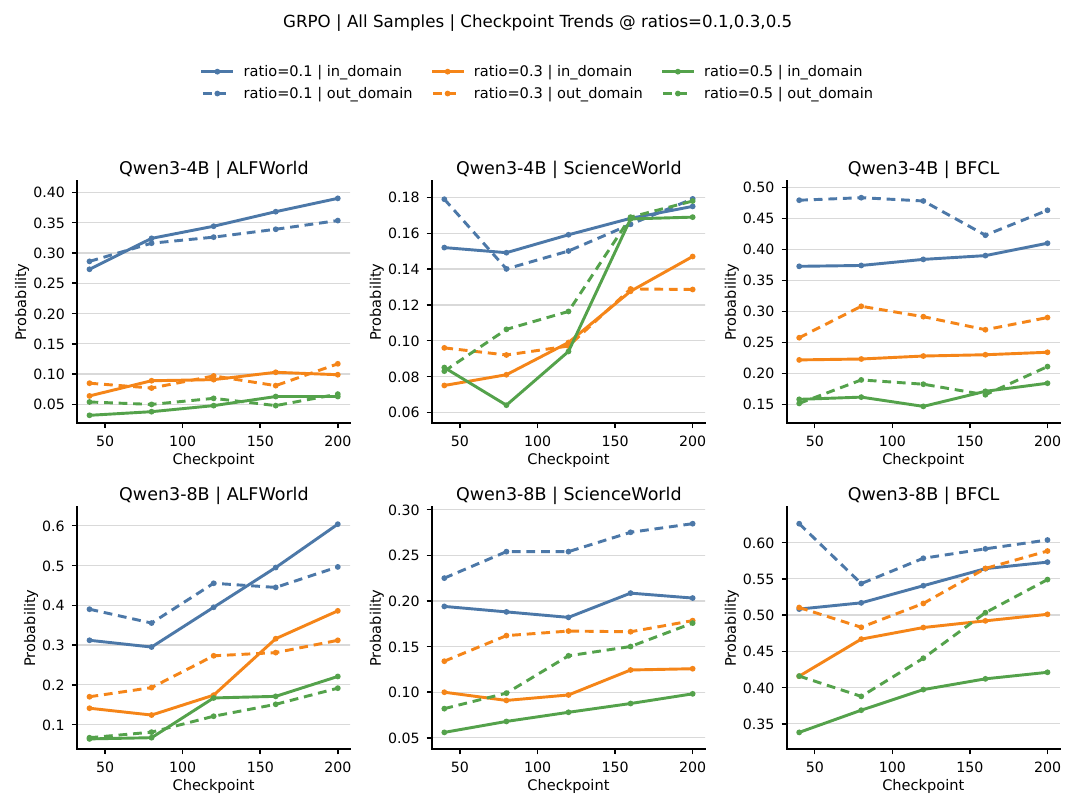}
\caption{Checkpoint-wise prompt-action consistency under perturbed reasoning traces during reinforcement learning.}\label{fig:grpo_all_samples_ckpt_nature}
\end{figure*}

\begin{figure*}[t]
\centering
\includegraphics[width=0.95\textwidth]{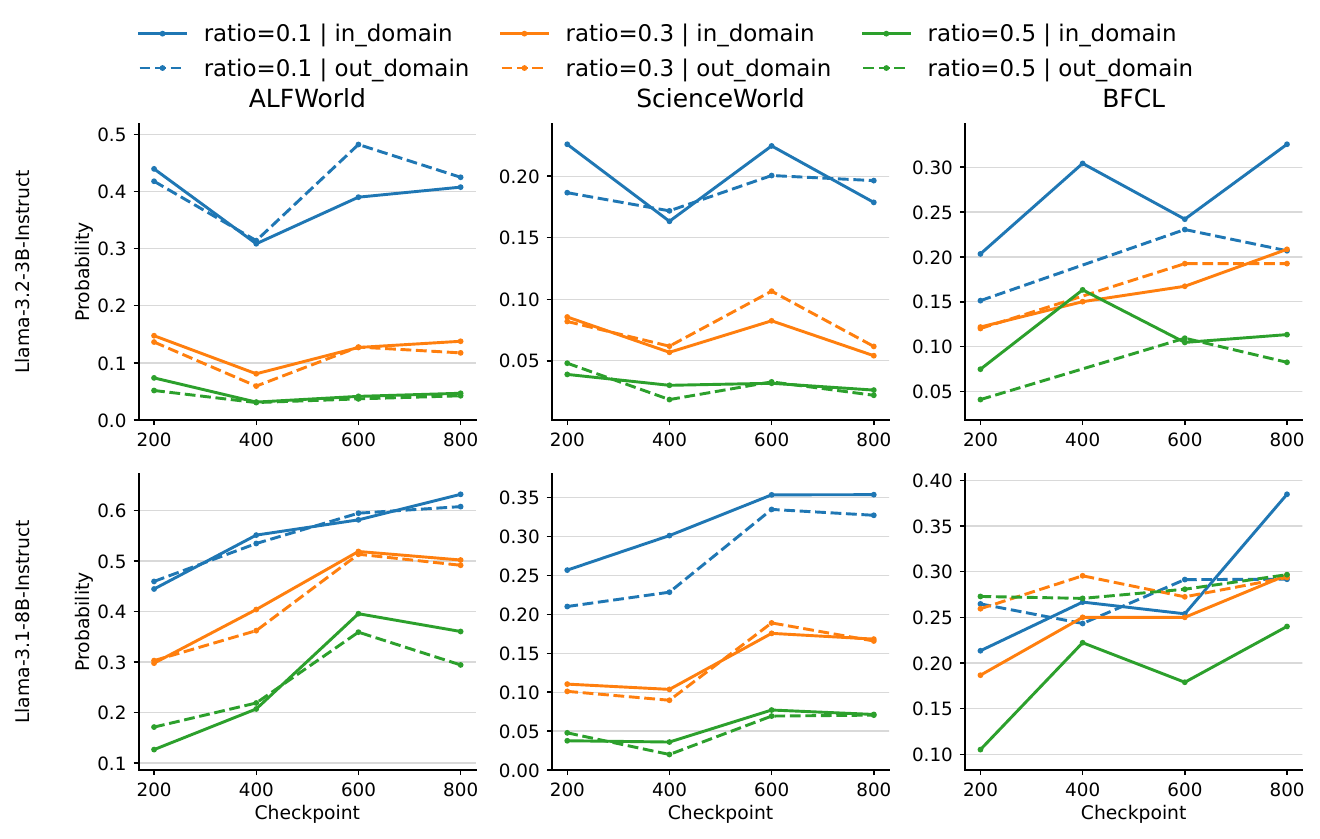}
\caption{Checkpoint-wise prompt-action consistency under perturbed reasoning traces for the Llama model.}\label{fig:all_samples_ckpt_nature}
\end{figure*}

\begin{figure}
\centering
\includegraphics[width=0.45\textwidth]{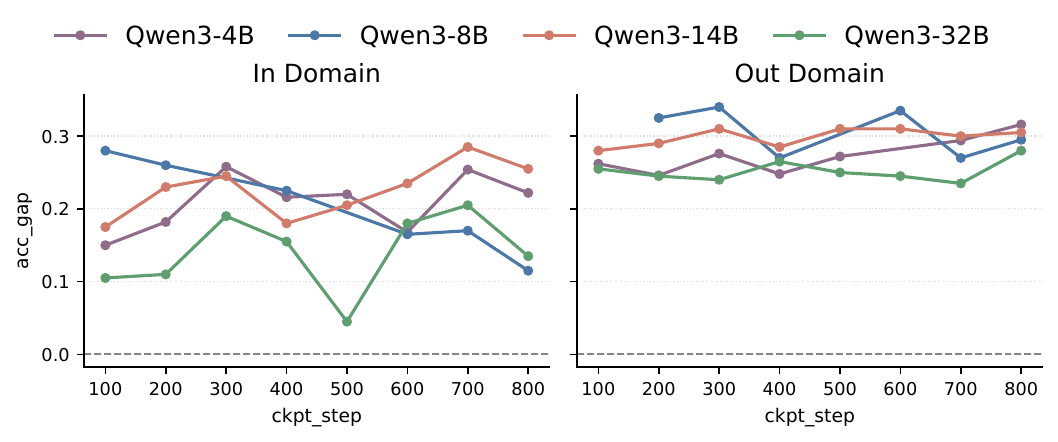}
\caption{Prompt/CoT-action success gap in BFCL.}\label{fig:bfcl_multi_model_acc_gap_plot}
\end{figure}

\begin{figure}
\centering
\includegraphics[width=0.45\textwidth]{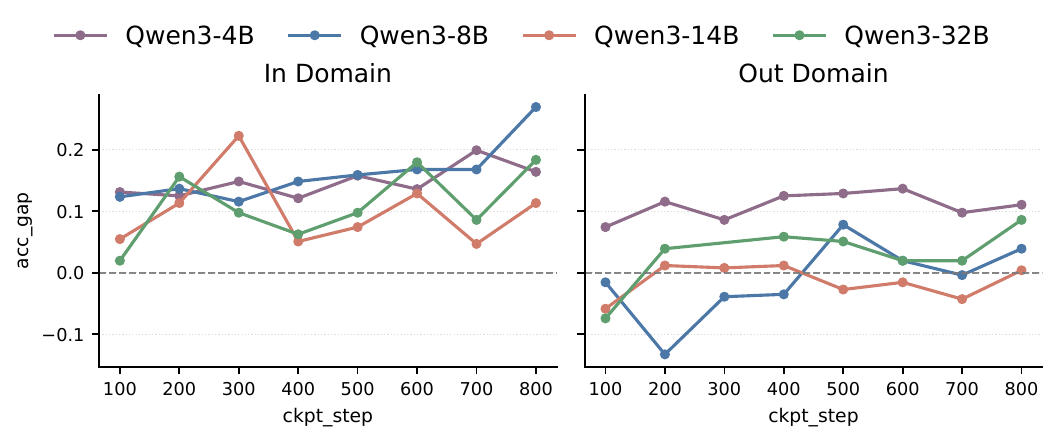}
\caption{Prompt/CoT-action success gap in ScienceWorld.}\label{fig:sci_multi_model_acc_gap_plot}
\end{figure}

\begin{figure*}[t]
\centering
\includegraphics[width=0.95\textwidth]{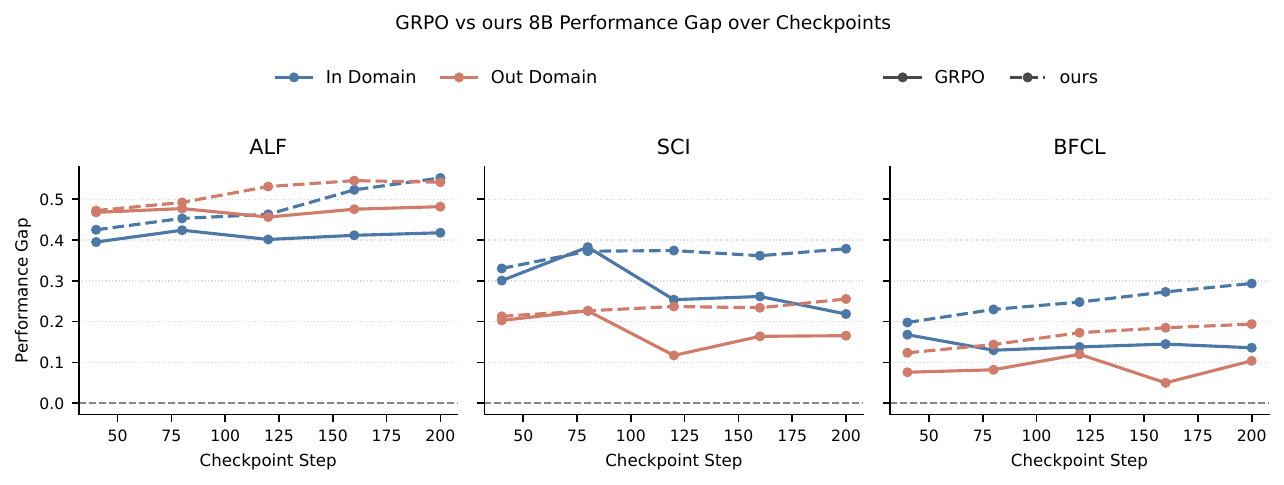}
\caption{Success gap between CoT actions and prompt actions under
reinforcement learning (GRPO).}\label{fig:grpo_vs_ours_8b_gap_three_task_plot}
\end{figure*}

\subsection{A Qualitative Example}
We conclude this section with one representative example, adapted from
the supplied experimental log, that illustrates how a coherent
rationale can still yield an action that does not materially improve
over the initial prompt-conditioned decision. In an ALFWorld task with
the goal ``put a cool tomato in the microwave,'' the interaction
history already records retrieving the tomato from the countertop. The
generated rationale nevertheless emphasizes examining the tomato again
to verify its temperature, even though the task context and recent
observations already constrain the next action substantially. This
example is consistent with the main-text diagnosis: a rationale may
remain locally plausible while still introducing unnecessary or
hallucinated intermediate justifications.

\section{Mechanistic Analysis}
We then move from behavioral patterns to a mechanistic account. The two
subsections below formalize why information from the prompt can dominate
action prediction and provide a gradient-based diagnostic that matches
the main-text interpretation.

\subsection{Value-Path Gradient Decomposition}\label{sec:value_path_derivation}
This section gives the full derivation behind the main-text claim that
larger attention mass on prompt tokens is associated with a larger
share of the optimization signal along the value path of
self-attention.

Consider one attention head at an action position $t$, with
\begin{equation}
q_t = W_Q h_t,\qquad
k_i = W_K h_i,\qquad
v_i = W_V h_i,
\end{equation}
\begin{equation}
\alpha_{t,i}
= \mathrm{softmax}\!\left(\frac{q_t^\top k_i}{\sqrt{d}}\right),
\qquad
o_t = \sum_{i \le t}\alpha_{t,i} v_i.
\end{equation}
If $\mathcal{L}_t = -\log p_\theta(a_t \mid x_{\le t})$ denotes the loss
for the action token at position $t$, then along the value path of
self-attention,
\begin{equation}
\frac{\partial \mathcal{L}_t}{\partial v_i}
= \alpha_{t,i}\,\delta_t,
\qquad
\delta_t = \frac{\partial \mathcal{L}_t}{\partial o_t},
\end{equation}
and therefore
\begin{equation}
\frac{\partial \mathcal{L}_t}{\partial W_V}
= \sum_{i \le t}\alpha_{t,i}\,\delta_t\,h_i^\top.
\end{equation}
Partitioning the available context into prompt tokens $P$ and CoT
tokens $C$ yields
\begin{equation}
\frac{\partial \mathcal{L}_t}{\partial W_V}
= \sum_{i \in P}\alpha_{t,i}\,\delta_t\,h_i^\top
+ \sum_{i \in C}\alpha_{t,i}\,\delta_t\,h_i^\top.
\end{equation}
This decomposition is the basis for the main-text interpretation: when
prompt tokens systematically receive larger attention weights during
action generation, they also receive a larger share of the
optimization signal along this pathway.

\subsection{Gradient-Based Diagnostic}
To complement this forward-pass view, we also compute a direct
gradient-based diagnostic of where the action-supervision signal flows
during backpropagation; details are provided here. We run a forward
pass on each sample and define an action-only loss
\begin{equation}
\mathcal{L}_{\mathrm{act}} = - \sum_{t=1}^{|y|} M(t)\log p_{\theta}(y_t \mid x_{<t}),
\end{equation}
where $M(t)=1$ only for positions in the final action span. This setup
asks a narrow question: when the model is trained to predict the final
action under teacher forcing, which part of the available context
receives the stronger update signal?

For each transformer block $\ell$, let $h_i^{(\ell)}$ denote the hidden
state at token position $i$. After backpropagating
$\mathcal{L}_{\mathrm{act}}$, we record the per-token gradient norm
\begin{equation}
g_i^{(\ell)} = \left\lVert \frac{\partial \mathcal{L}_{\mathrm{act}}}{\partial h_i^{(\ell)}} \right\rVert_2.
\end{equation}
Let $P$ be the set of prompt tokens and $C$ the set of CoT tokens. We
measure the gradient share on prompt tokens
\begin{equation}
R_{\mathrm{prompt}}^{(\ell)} =
\frac{\sum_{i \in P} g_i^{(\ell)}}{\sum_{i \in P \cup C} g_i^{(\ell)}}.
\end{equation}

\begin{table*}[htbp]
  \centering

    \begin{tabular}{ccccc}
    \toprule
          &       & Qwen3-4B & Qwen3-8B & Qwen3-14B \\
    \midrule
    \multirow{2}[2]{*}{ALFWorld} & prompt & 73.2  & 72.7  & 73.5 \\
          & response & 26.8  & 27.3  & 26.5 \\
    \midrule
    \multirow{2}[2]{*}{ScienceWorld} & prompt & 76.5  & 78.9  & 79.2 \\
          & response & 23.5  & 21.1  & 20.8 \\
    \midrule
    \multirow{2}[2]{*}{BFCL} & prompt & 84.4  & 84.6  & 82.7 \\
          & response & 15.6  & 15.4  & 17.3 \\
    \bottomrule
    \end{tabular}%
    \caption{Action-only gradient mass is concentrated on prompt tokens rather than CoT tokens across environments and model sizes. Entries report the share of per-token gradient norm assigned to prompt versus response tokens under the action-only loss.}\label{tab:per_token_gradient_norm}
\end{table*}%

\begin{figure*}[t]
\centering
\includegraphics[width=0.95\textwidth]{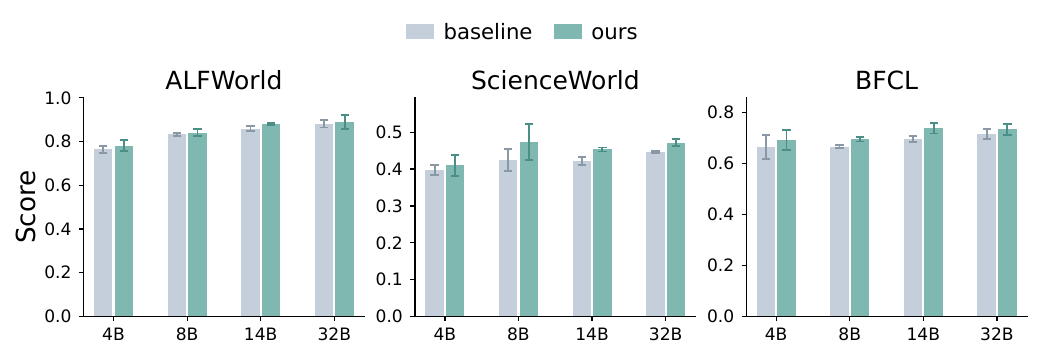}
\caption{Mean evaluation score on in-domain tasks under vanilla CoT supervision
(\textit{baseline}) and reduced action supervision (\textit{ours})
across environments and model sizes. Reduced action supervision
improves performance in most environment-model combinations, with the
largest gains appearing in ALFWorld and ScienceWorld.}\label{fig:per_cot_id_nature}
\end{figure*}

As reported in Table~\ref{tab:per_token_gradient_norm}, the action-only
gradient is substantially more concentrated on prompt tokens than on
reasoning tokens across all four models and three environments. This
is consistent with a systematic optimization bias: standard action
supervision drives updates predominantly through the prompt-based
pathway rather than through the rationale.

Together, the attention and gradient evidence offer a structural
explanation compatible with the main-text interpretation. Agent
prompts are substantially longer than the reasoning traces they
surround, and this length asymmetry translates into an optimization
asymmetry: the prompt side receives the dominant share of attention
and gradient signal during action prediction. Standard CoT
supervision therefore appears to disproportionately strengthen the
direct prompt-to-action pathway. This mechanistic pattern matches the
behavioral diagnosis that later checkpoints make the final action more
predictable from the prompt and more resistant to shortcut conflict.

\section{Prompts}
Finally, we provide the full prompts used for training and evaluation,
including the environment prompts and the action-comparison prompt used
for local judgments.

\begin{lstlisting}[
  style=promptstyle,
  caption={Prompt for ALFWorld.},
  label={lst:prompt_alfworld}
]
You are an expert agent operating in the ALFRED Embodied Environment.
Your task is to: {task_description}

Prior to this step, you have already taken {step_count} step(s).
Below are the most recent {history_length} observations and the
corresponding actions you took:
{action_history}

You are now at step {current_step} and your current observation is:
{current_observation}

Your admissible actions of the current situation are: [{admissible_actions}].

Now it's your turn to take an action.
You should first reason step-by-step based on the guideline about the
current situation. This reasoning process MUST be enclosed within
<thinking> </thinking> tags.
Once you've finished your reasoning, choose an admissible action for
the current step and present it within <action> </action> tags.
\end{lstlisting}

\begin{lstlisting}[
  style=promptstyle,
  caption={Prompt for ScienceWorld.},
  label={lst:prompt_scienceworld}
]

You are an expert agent operating in the ScienceWorld environment,
which is a text-based virtual environment centered around accomplishing
tasks from the elementary science curriculum.
Your current task is: {task_description}

Prior to this step, you have already taken {step_count} step(s).
Below are the most recent {history_length} observations and the
corresponding actions you took:
{action_history}

You are now at step {current_step} and your current observation is:
{current_observation}

Here are the actions you may take:
{action_descriptions}

Current available actions:
{available_actions}

Now it's your turn to take an action.
You should first reason step-by-step about the current situation.
This reasoning process MUST be enclosed within <thinking> </thinking>
tags.
Once you've finished your reasoning, choose an appropriate action for
the current step and present it within <action> </action> tags.

\end{lstlisting}
For BFCL, we directly adopt the prompt from \citet{agentevolver}.

\begin{lstlisting}[
  style=promptstyle,
  caption={Prompt to judge which action is better.},
  label={lst:prompt_action_judge}
]
You are a strict evaluator of Agent action quality.
Please judge which of the two actions is better based only on the given original user prompt.
The better action should better match the task goal, better fit the current observation, be more executable, be safer, and be more likely to advance task completion.
If the two actions are of similar quality, judge them as similar.

Original user prompt:
{user_prompt}

Candidate action A:
{cot_action}

Candidate action B:
{direct_action}

Please think step by step first, and then output your final answer inside <judge></judge> tags.
Inside the tag, output exactly one of the following three labels:
- A_better
- B_better
- similar
\end{lstlisting}

\end{document}